\documentclass[letterpaper, 10 pt, conference]{ieeeconf}  

\usepackage{graphicx} 
\usepackage{csvsimple}
\usepackage{xcolor}
\usepackage{pdfpages}
\usepackage{array} 
\usepackage{booktabs} 
\usepackage{multirow} 
\usepackage{tabularx} 
\usepackage{caption} 
\usepackage{float} 
\usepackage{amsmath}  
\usepackage{subcaption}  

\usepackage[backend=biber,style=ieee]{biblatex} 
\IEEEoverridecommandlockouts                              

\title{\LARGE \bf
The impact of Compositionality in Zero-shot Multi-label action recognition for Object-based tasks
}

\author{Carmela Calabrese$^{*}$, Stefano Berti$^{*}$, Giulia Pasquale, Lorenzo Natale
\\ 
Humanoid Sensing and Perception, Italian Institute of Technology, Genoa, 16163, Italy
\\
\textit{name.surname@iit.it}
\thanks{* These authors equally contributed}
}

\begin{document}

\maketitle
\thispagestyle{empty}
\pagestyle{empty}

\begin{abstract}
Addressing multi-label action recognition in videos represents a significant challenge for robotic applications in dynamic environments, especially when the robot is required to cooperate with humans in tasks that involve objects. Existing methods still struggle to recognize unseen actions or require extensive training data. To overcome these problems, we propose Dual-VCLIP, a unified approach for zero-shot multi-label action recognition. Dual-VCLIP enhances VCLIP, a zero-shot action recognition method, with the DualCoOp method for multi-label image classification. The strength of our method is that at training time it only learns two prompts, and it is therefore much simpler than other methods. We validate our method on the Charades dataset that includes a majority of object-based actions, demonstrating that -- despite its simplicity -- our method performs favorably with respect to existing methods on the complete dataset, and promising performance when tested on unseen actions. Our contribution emphasizes the impact of verb-object class-splits during robots' training for new cooperative tasks, highlighting the influence on the performance and giving insights into mitigating biases. 

\end{abstract}

\section{INTRODUCTION}
Addressing multi-label action recognition in videos represents a significant challenge for many real-world robotic applications.
Robots play a key role in modern workplaces where they are often used for repetitive tasks.
To cooperate with humans, robots are expected to actively interact in dynamic and poorly structured environments, where a priori definition of objects and actions is non-trivial \cite{ayub2020tell}. 

\begin{figure}[h]
    \centering
    \includegraphics[width=0.45\textwidth]{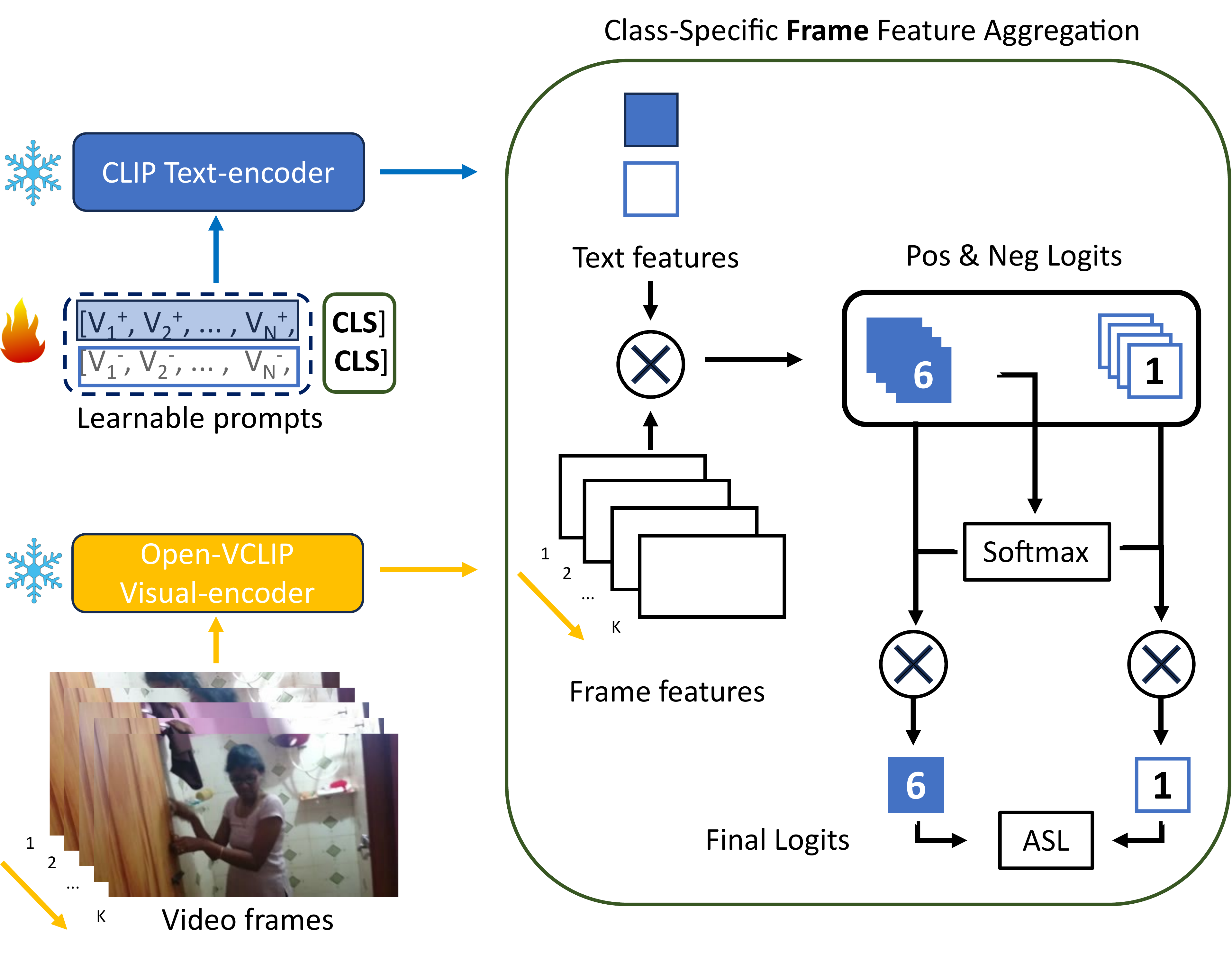}
    \captionsetup{font=small} 
    \caption{\textbf{Overview of our proposed Dual-VCLIP for multi-modal multi-label action recognition.} It has two main components: a video encoder, a textual encoder. DualCoOp learns a pair of positive and negative prompts to quickly adapt pretrained vision-text encoders to the Multi Label Recognition (MLR) task. For each class, two prompts generate two contrastive (positive and negative) textual embeddings as the input to the text encoder. Furthermore, we propose Class-Specific Frame Feature Aggregation to first project each frame’s feature to the textual space and then aggregate the temporal logits by the magnitude of class-specific semantic responses. During training, we apply the asymmetric loss from \cite{sun2022dualcoop} to optimize learnable prompts while keeping other network components frozen.}
    \label{fig:architecture}
\end{figure}

Most recognition methods result impractical for these purposes since they require that all target classes are known beforehand, and an extensive number of examples is provided for each class at training time \cite{chiatti2020task}. For this reason, in many applications, robots need to be incrementally trained to recognize the specific objects \cite{chiatti2020task, ayub2020tell} or actions \cite{wang2021demograsp} needed for a specific joint task, using only limited set of visual demonstrations provided by the final user.
Conventional action recognition methods often focus on learning the visual modality alone, neglecting additional information sources.
Recent works leverage large-scale pretrained foundation models due to their adaptability to transfer knowledge on a different set of downstream tasks.
Through natural language processing and multimodal information, these methods learn an alignment between textual and visual spaces to compensate for limited labeling. 
The generalization capability of CLIP-based architectures becomes fundamental for robotics scenarios where training on much smaller datasets while maintaining nearly the same performances becomes desirable \cite{chiatti2020task, ayub2020tell, wang2021demograsp, berti2022one}.

While Contrastive Language-Image Pretraining (CLIP) \cite{radford2021learning} has demonstrated impressive zero-shot learning abilities for image understanding, there has been yet limited exploration of CLIP's potential for zero-shot video recognition. To extend image-text models to video, recent work has focused on the visual domain by aggregating temporal information with new modules \cite{adeli2015multi, wang2021actionclip}. In addition, most recent action recognition methods focus on identifying single activities, assuming only one action per video-clip and discarding the potential simultaneous occurrence of simultaneous actions \cite{lentzas2022multilabel}, that can be reductive and damaging for an efficient HRI.
Instead, robots need to be incrementally trained to recognize the multiple and specific objects \cite{chiatti2020task, ayub2020tell} or actions \cite{wang2021demograsp} needed for a specific joint task, using only limited set of visual demonstrations provided by the final user \cite{berti2022one}, leveraging the adaptability of foundation models to transfer knowledge to different downstream tasks. In this perspective, Compositional Zero-Shot Learning tries to imitate the human ability to reduce complex stimuli to primitives, learn by discrimination, and generalize to recognize representations of unseen compositions composed of seen concepts \cite{sucholutsky2024using}  based on prototype-based categorization (as opposed to exemplar-based) for categorical inferences \cite{malaviya2022can}.

To overcome these limitations, we propose a new approach called Dual-VCLIP which offers a unified solution for different aspects in zero-shot multi-labelling action recognition tasks, and we investigate preliminary results on our solution for multi-labelling action recognition. 
In this paper, the proposed method, Dual-VCLIP, is an efficient and minimalist approach that extends the zero-shot multi-labelling object recognition method DualCoOp \cite{sun2022dualcoop} to action recognition in video through OpenVCLIP \cite{weng2023open}, a simple yet effective approach that transforms CLIP into a strong zero-shot video classifier that can recognize unseen actions at inference time. 

Figure \ref{fig:architecture} shows an overview of our proposed model.
Preliminary results suggest that this framework can achieve competitive results on zero-shot action recognition on Charades \cite{sigurdsson2016hollywood}. 

Our contribution is summarized as follows: (1) we propose a new method for zero-shot multi-label action recognition, which remains effective while trained only small set of parameters, and compares favorably against existing methods (2) we stress on the relevance of class-splits based on verb-object pairs to analyze bias in the temporal or spatial dimension. For the sake of reproducibility, dataset splits and code will be made publicly available upon acceptance.

\section{Related work}

\subsection{Multi-label action recognition}
Recognizing the behaviors of humans in videos is crucial for Human-Robot Interaction (HRI) across different domains, such as services \cite{anagnostis2021human}, industrial \cite{roitberg2014human}, healthcare \cite{rodomagoulakis2016multimodal}. Although most recent action recognition methods focused on identifying single activities, assuming only one action per video-clip and discarding the potential simultaneous occurence of simoultanoues actions \cite{lentzas2022multilabel}, can be reductive and damaging for an efficient HRI. Many existing approaches primarily concentrate on single-activity recognition and utilize sigmoid as the output activation function to provide multi-label predictions \cite{wang2021actionclip}. Some methods solved multi-label recognition at the feature level, by extracting independent feature descriptors for each activity instead of shared feature representations, and thus capturing correlations among different actions \cite{zhang2021multi}. Weakly-supervised learning approaches \cite{adeli2015multi} successfully advanced multi-label learning for video analysis, localizing spatio-temporal discriminative regions inferred by clustering regions of common texture and motion features.

\subsection{Multi-label zero-shot action recognition}
Robots are expected to actively interact in dynamic and poorly structured environments, where a priori definition of objects and actions is non-trivial \cite{ayub2020tell}. Most recognition methods are impractical for these purposes as they require prior knowledge of all target classes and extensive training examples for each class \cite{chiatti2020task}. For this reason, in many applications, robots need to be incrementally trained to recognize the specific objects \cite{chiatti2020task, ayub2020tell} or actions \cite{wang2021demograsp} needed for a specific joint task, using only limited set of visual demonstrations provided by the final user \cite{berti2022one}, leveraging the adaptability of foundation models to transfer knowledge to different downstream tasks. In this perspective, zero-shot recognition aims to classify novel categories during testing. However, multi-label zero-shot action recognition remains relatively unexplored. Kerrigan, et al. \cite{kerrigan2021reformulating} reformulated multi-label zero-shot action recognition in terms of pairwise scoring function between confidence scores for each class independently learnt, showing strong performances in the prediction of semantically distinct classes. The same can be afforded exploring relationships between multiple labels with graph structure to learn semantic dependencies between words in a class name and leveraging information propagation mechanism from the semantic label space to predict unseen class labels \cite{lee2018multi}. Wang and Chen \cite{wang2020multi} achieved multi-label zero-shot recognition by measuring relatedness scores of action labels in the joint latent visual and semantic embedding spaces. Other approaches consider losses with evidence debiasing constraints to reduce the static bias of video representations, via shared multi-attention model to map only relevant segments of an image to a join visual-label embedding space \cite{huynh2020shared} or via evidential neural networks that estimate multi-action uncertainty to detect novel actions\cite{zhao2023open}.

It is worth mentioning that the methods mentioned above do not adapt the foundation model to the distribution of the dataset by looking at a portion of the classes.

\subsection{Compositionality in action recognition}
Human activities involve a series of interactions occurring over time and with various objects in space \cite{ji2020action, baradel2018object}. This is particularly crucial in Human-Robot interaction scenarios where tasks often involve specific objects. However, existing action recognition methods still struggle with novel combinations of known actions and objects \cite{materzynska2020something}, posing challenges for robotic applications. Enhancing current models with compositional generalization could be a crucial and promising avenue for improving action understanding. The work of Sigurdsson et al. \cite{sigurdsson2017actions} provides a comprehensive analysis to identify the key factors contributing to significant improvements in action understanding, emphasizing the role of objects, verbs, and sequential reasoning, providing fine-grained understanding of objects when combined with temporal reasoning. Their analysis sheds light on specific training methodologies, like avoiding treating actions as singular events occurring within videos, leveraging hierarchical event decomposition to facilitate few-shot action recognition \cite{ji2020action}. For instance, Materzynska et al. \cite{materzynska2020something} introduced a compositional framework wherein there is no overlap between the verb-noun combinations in the training and testing datasets. This compositional approach holds significant promise for the multi-label fine-grained video classification task 

\begin{figure*}[t]
  \centering
  \vspace{0.5cm}
  \begin{subfigure}[b]{0.49\textwidth}
    \centering
    \includegraphics[width=\textwidth]{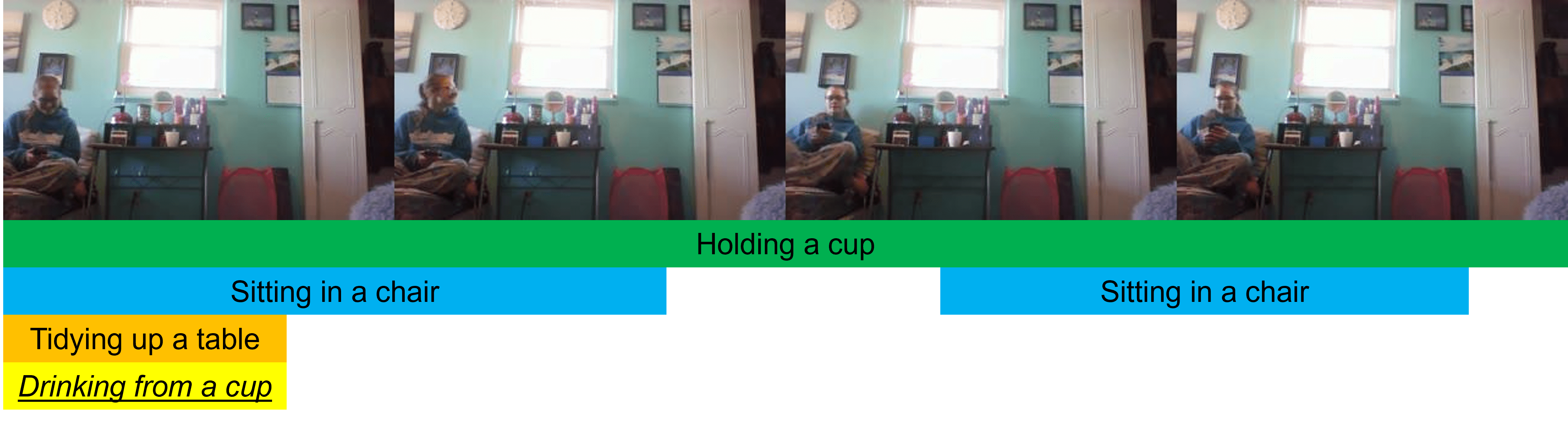}
    \label{fig:subfig1}
    \vspace{-0.7\baselineskip} 
  \end{subfigure}
  \hfill
  \begin{subfigure}[b]{0.49\textwidth}
    \centering
    \includegraphics[width=\textwidth]{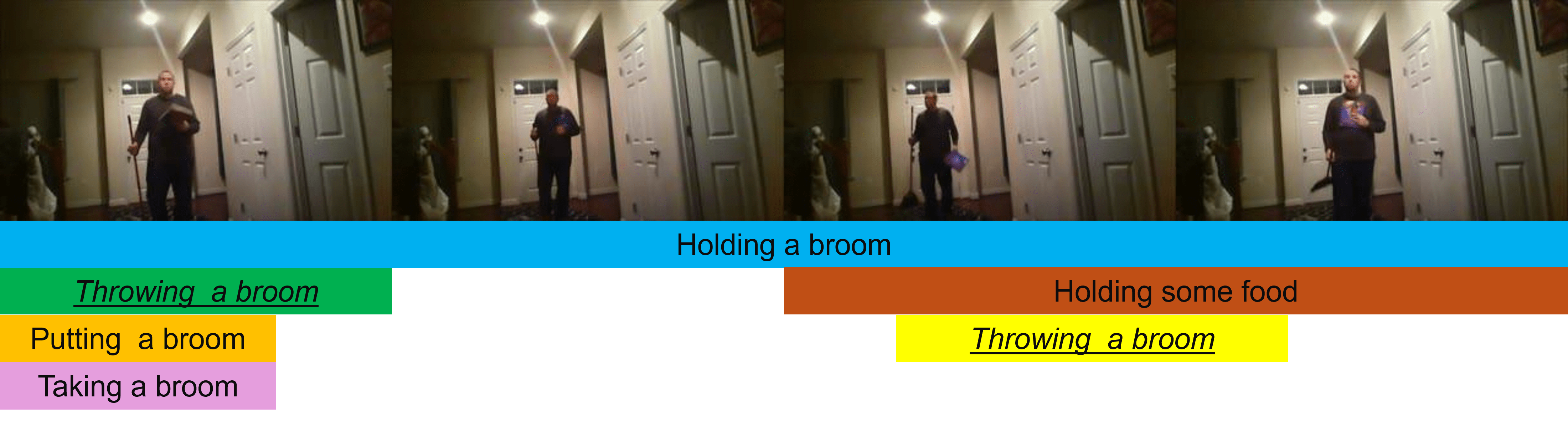}
    \label{fig:subfig2}
    \vspace{-0.7\baselineskip} 
  \end{subfigure}
  \captionsetup{font=small} 
  \caption{\textbf{Inference per-frame with our method Dual-VCLIP on two examples from the Charades' test set.} Underlined classes are unseen- they did not appear at training time. Classes that appears only in one frame are discarded.}
  \label{fig:entirefig}
\end{figure*}
\vspace{-0.7\baselineskip}

\section{Background}
Vision-language models pretrained on large-scale datasets such as CLIP \cite{radford2021learning} align the feature of visual and textual representations, thus allowing for zero-shot image classification.
Formally, given an image $x$ and a set of classes as textual descriptions $y$, the probability for each class $y_j \in y$ is:

\begin{equation}
    P(y_j|x) = \frac{\text{exp}(\text{sim}(\text{CLIP}_V(x), \text{CLIP}_T(y_i)))}{\sum\limits_j \text{exp}(\text{sim}(\text{CLIP}_V(x), \text{CLIP}_T(y_j)))}
\end{equation}
where $\text{CLIP}_V$ is the visual encoder of $\text{CLIP}$, $\text{CLIP}_T$ is the textual encoder of $\text{CLIP}$, $\text{``sim''}$ is a similarity function such as the cosine similarity and ``exp'' is the exponential function.

The original image CLIP model does not have the ability to aggregate temporal relations among video frames and thus is not suitable to identify actions.
OpenVCLIP \cite{weng2023open} adapts CLIP framework to sequences of images without additional parameters, but just by expanding the temporal attention view for every self-attention layer.
Formally, the self-attention layers of CLIP are computed as follows:

\begin{equation}
    y_{t} = \text{Softmax}\left(\frac{q_{t}K_{t}^{T}}{\sqrt{d}}\right)V_{t}
\label{softmax_clip}
\end{equation}

and the modified version for OpenVCLIP as follows:

\begin{equation}
    y_{t} = \text{Softmax}\left(\frac{q_{t}K_{(t-1)\sim(t+1)}^{T}}{\sqrt{d}}\right)V_{(t-1)\sim(t+1)}
\label{softmax_openvclip}
\end{equation}

where $d$ refers to the dimension of the model, $q_{t}$ refers to the query vector of the $t$-th frame, $K_t^T$ is the transpose of the matrix composed of key vectors and $V_t$ is the matrix composed of value vectors in the $t$-th frame.
The global temporal information is obtained from the stacking of the self-attention layers.
The resulting architecture is then fine-tuned on a large-scale action recognition dataset such as Kinetics-400 \cite{kay2017kinetics}.
The resulting weights are averaged with the original CLIP weights using a technique named Interpolation Weights Optimization \cite{ilharco2022patching}, that allows a model to learn new concepts while preserving original knowledge.

With CLIP we obtain a similarity matrix between images and textual descriptions, that can be used to obtain the probability for a class in a certain image with a softmax function, like in Eq. (\ref{softmax_clip}).
Anyway, the softmax is not designed for the multi-label classification, and a threshold on the logits is not possible because each class may need a different value, decreasing the performance on multi-label zero-shot classification.
DualCoOp \cite{sun2022dualcoop} proposes to learn a positive and a negative prompt and to consider a class present in an image only if the similarity between the image and the positive prompt is higher that the similarity between the image and the negative prompt.
In the supervised setting, a positive and a negative prompts are learned for each class, while in the zero-shot setting a positive and a negative prompts are learned and given to the textual encoder together with the name of the class.
Since a single image can contain different classes, the feature of that image is required to contain information about different classes in different portion of the image, thus obtaining sub-optimal performance since spatial information is reduced and different classes representations are mixed.
To overcome this issue, \cite{sun2022dualcoop} reformulate the last multi-headed attention pooling layer of the visual encoder in CLIP: they do not pool the visual attention map, but they project the visual feature of each region to the textual space.
This allows to create class-specific region features, that can correctly represent different classes in different portion of the image.

\section{Problem definition}
In zero-shot setting, the model is trained on a source dataset $D_{train}$ with classes $Y_{train}$ and tested directly on a target dataset $D_{test}$ with classes $Y_{test}$, where \mbox{$Y_{train} \cap Y_{test} = \emptyset$}.

We also consider the setting where $Y_{train}$ and $Y_{test}$ are part of the same dataset and the name of the classes have the form \textit{subject + object + verb}.
This allow to split the classes in a different way, for example by masking out certain objects or verbs at training time to see how this affects the performances of the model and to understand its possible temporal or spatial biases.

\section{Method}
In this paper, we introduce a novel method for zero-shot multi-label action recognition, named Dual-VCLIP.
Our method learns a prompt that adapt the knowledge of the foundation model to the distribution of the target dataset, while exhibiting competitive performance on both seen and unseen classes.
Dual-VCLIP, as DualCoOp \cite{sun2022dualcoop}, learns a pair of “prompt” contexts as two learnable sequences of word vectors, to provide positive and negative contexts for a given category name $y$.
In Figure \ref{fig:architecture}, we provide a schematic overview of our method.

Formally, to check if an action $y$ is contained in a video $x$, we extract the features of the video $x$ with OpenVCLIP as \mbox{$v = \text{OpenVCLIP}(x)$}. Then we create the positive and negative class-prompts \mbox{$\hat{y}_j^+ = [P^+, y_j]$} and \mbox{$\hat{y}_j^- = [P^-, y_j]$}, that are the concatenation of the learnable positive/negative prompts $P^+$ and $P^-$ and the class name $y_j$.
Then we encode the class-prompts with the textual encoder of CLIP:
\begin{equation}
    \hat{t}_{j}^{+} = \text{CLIP}_T(\hat{y}_{j}^{+})  
\end{equation}
\begin{equation}
    \hat{t}_{j}^{-} = \text{CLIP}_T(\hat{y}_{j}^{-})  
\end{equation}
and we compute the similarity between the feature extracted from the video and the positive and negative encoded class-prompts:
\begin{equation}
   S_{j, t}^+ = \text{sim}(V, \hat{t}_{j}^{+})  
\end{equation}
\begin{equation}
   S_{j, t}^- = \text{sim}(V, \hat{t}_{j}^{-})  
\end{equation}
where $t$ indicates the $t$-th frame of the sequence and ``sim'' is a similarity function, e.g., the cosine similarity.

Finally, for each class, we perform aggregation of all temporal logits, in which the weight for each logit is determined by its relative magnitude:
\begin{equation}
   S_j^+ = \sum_t (\text{Softmax}(S_{j, t}^+) \times S_{j, t}^+)  
\end{equation}
\begin{equation}
   S_j^- = \sum_t (\text{Softmax}(S_{j, t}^+) \times S_{j, t}^-)  
\end{equation}
We call this processing as \textit{Class-Specific Frame Feature Aggregation}.
It can be noted that this process is similar to the one of DualCoOp, but we exploit the temporal dimension instead of the spatial dimension.
At training time, the similarities are given to an asymmetric loss module, as in DualCoOp \cite{sun2022dualcoop}, while at inference time a class $j$ is predicted if $S_j^+ > S_j^-$.
We freeze the weights of all the components except for the learnable prompts, that has a reduced number of parameters (64x512 in our experiments).

\subsection{Experiments}

\textbf{Dataset.} We conduct experiments on Charades dataset \cite{sigurdsson2016hollywood} to evaluate zero-shot multi-label action recognition. It contains 66,50k annotations for 157 actions and it is divided into 7985 and 1863 video-clips for training and testing sets, respectively. This dataset is challenging for multi-label action recognition, since it contains clips from hundreds of people recording video-based human activity in their own home. The average duration of video-clips is $~30s$ and an average number of actions involved in a video-clip is 6.8. Since the public split is thought on an instance-level (all the 157 classes appears in both training and testing set) and cannot be directly applied to zero-shot tasks, we define three splits at the class-level. In other words, we define which of the 157 classes the model sees from the training split at training time, such that at testing time we have that the model has never seen a certain number of classes.

We define three different splits for the purposes of this work:
\begin{itemize}
\item Random Split: we randomly choose half of the classes in the test dataset, irrespective of a fair balance between verbs and objects. Specifically, for this experimental setting, we consider $Y_{train}$ and $Y_{test}$ respectively contain $50\%$ and $50\%$ of the total number of classes in the dataset.
\item Verbs Split: we first cluster classes according to the verb in the label $C_{verb}$ and then randomly choose $50\%$ of the clusters $C_{verb}$. All the classes where those verbs appear, belong to the test set $Y_{test}$, the remaining to the training set $Y_{train}$.
\item Objects split:  we first cluster classes according to the object in the label $C_{obj}$ and then randomly choose $50\%$ of the clusters $C_{obj}$. All the classes where those objects appear, belong to the test set $Y_{test}$, the remaining to the training set $Y_{train}$.
\end{itemize}
Table~\ref{tab:split_description} reports splits' details on the following evaluation protocols.

\begin{table}[htbp]
    \centering
    \begin{tabularx}{0.48\textwidth}{lXXc} 
        \toprule
         & \textit{Unseen Verbs} & \textit{Unseen Objects} & \textit{$\#$ Unseen Classes}\\
        \midrule
        Random Split & $26.3\%$ (10/38) &  $8.1\%$ (3/37) & 79\\
        Verb Split &$47.3\%$ (18/38) & $5.4\%$ (2/37) & 84\\
        Object Split & $34.2\%$ (13/38) & $48.6\%$ (18/37) & 82\\
        \bottomrule
    \end{tabularx}
    \captionsetup{font=small} 
    \caption{\textbf{Splits' Details.} Verbs and Objects in Testing Classes split into Seen (appeared during training phase) and Unseen (did not appear in training phase). About $50\%$ of classes are used for zero-shot.}
    \label{tab:split_description}
\end{table}

\textbf{Evaluation protocols.}
We evaluate all methods with both zero-shot setting (only $50\%$ of seen classes) and on all the entire Charades dataset.
As done in previous works \cite{sun2022dualcoop}, we report the mAP value on the test set for each split.

\textbf{Implementation.}  We use the ViT-B-16 version of CLIP for all the experiments.
The OpenVCLIP model is obtained from a weight interpolation with the original CLIP's weights using a patching ratio of 0.5.
The model is trained on 4 TeslaV100 GPUs with a total batch size of 64 until convergence, with one warm-up epoch.
We use SGD as optimizer, with a learning rate that starts from 1e-3 and follows a cosine annealing scheduling.
We set the number of context vectors to 64.
We used 16 frames for inference and training with a resolution of 224.

\textbf{Baseline.} To evaluate the effectiveness of Dual-VCLIP in the zero-shot setting, we compare it with the following baselines: (1) CLIP-Hitchiker \cite{bain2022clip}, (2) VideoCOCA \cite{yan2022videococa}, (3) OpenVCLIP \cite{weng2023open} and (4) MSQnet \cite{mondal2023actor}.
CLIP-Hitchiker and VideoCOCA are two methods that are directly tested on the test set of charades without any additional training.
We also tested MSQNet using the code released by the authors. It is a method for zero-shot action recognition that requires to be trained on part of the classes, similarly to our method. Unfortunately, we were not be able to reproduce the performance reported in their work, while achieving consistently lower results on our train-test. We believe this may be due to the fact that we could not find the details of the train-test split they used in their experiment, but further investigation is needed to confirm these experiments. 

Moreover, we compare with OpenVCLIP adapted to multi-labelling, where we use a threshold of 0.5 on the similarity scores between videos and classes instead of a softmax over all classes.

\section{Results}
\subsection{Comparison with SOTA}
In Table \ref{tab:comparison} we report the mAP results in zero-shot- when only $50\%$ of classes has been seen during training- and on the full Charades testing set, for the different action recognition methods. The results show that our method has lower performance with respect to the competitor MSQNet. However, these results are not comparable since training the two methods on their splits was not possible, while achieving consistently lower results with MSQNet on our train-test. Instead, when homogeneous comparison is possible (on the full testing dataset), it is remarkable that our method returns competitive mAP values with respect to the SOTA methods . This shows the convenience of adding a learnable context vector to improve the performance in generalized setting (compared to OpenVCLIP). It is noteworthy that our method learns a limited number of parameters with respect to other methods- we learn just two prompts that have 64 context tokens each.
 
\begin{table}[htbp]
    \centering
    \begin{tabularx}{0.4\textwidth}{l|X|X} 
        \toprule
        Method & mAP \textit{ZSL} ($50\%$ Seen classes) & mAP on full Charades testing set \\
        \midrule
        CLIP-Hitchhiker$^{*}$ \cite{bain2022clip} & - & 21.1 \\
        VideoCoCa$^{*}$ \cite{yan2022videococa} & - & 25.8 \\
        MSQNet$^{**}$ & 30.9 & - \\
        OpenVCLIP adapted & - & 18.82 \\
        \midrule
        Dual-VCLIP (ours) & 16.65 & 22.98 \\
        \bottomrule
    \end{tabularx}
    \captionsetup{font=small} 
    \caption{\textbf{Comparison of different zero-shot action recognition methods on Charades} ($^{*,**}$ values reported from literature:$^{*}$ unavailable code, $^{**}$ unavailable splits).}
    \label{tab:comparison}
\end{table}

\subsection{Compositionality}
To implement this approach effectively in Human-Robot interaction scenarios, it is crucial to ensure its capability in fine-grained action recognition. This entails the ability to discern between actions that involve the same object, such as ``passing a box'', ``looking at a box,'' and ``opening a box.'' The Charades dataset serves as a suitable resource for conducting this detailed analysis. Notably, the Charades dataset comprises 157 action classes, with categories that extend beyond atomic actions. Specifically, 146 classes can be deconstructed into ``Verb+Object'' combinations, while 11 classes involve only a verb. Upon scrutinizing the semantics of these classes, we identified 38 verbs and 37 objects appearing in various combinations. To clarify the knowledge acquired by our method, we calculated the mean Average Precision (mAP) across five distinct subsets of the testing set of the Random Split. These subsets distinguish between scenarios where verbs or objects are either recognized or remain unseen.

\begin{figure}[t]
  \vspace{0.2cm}
\centering
\includegraphics[width=0.95\columnwidth]{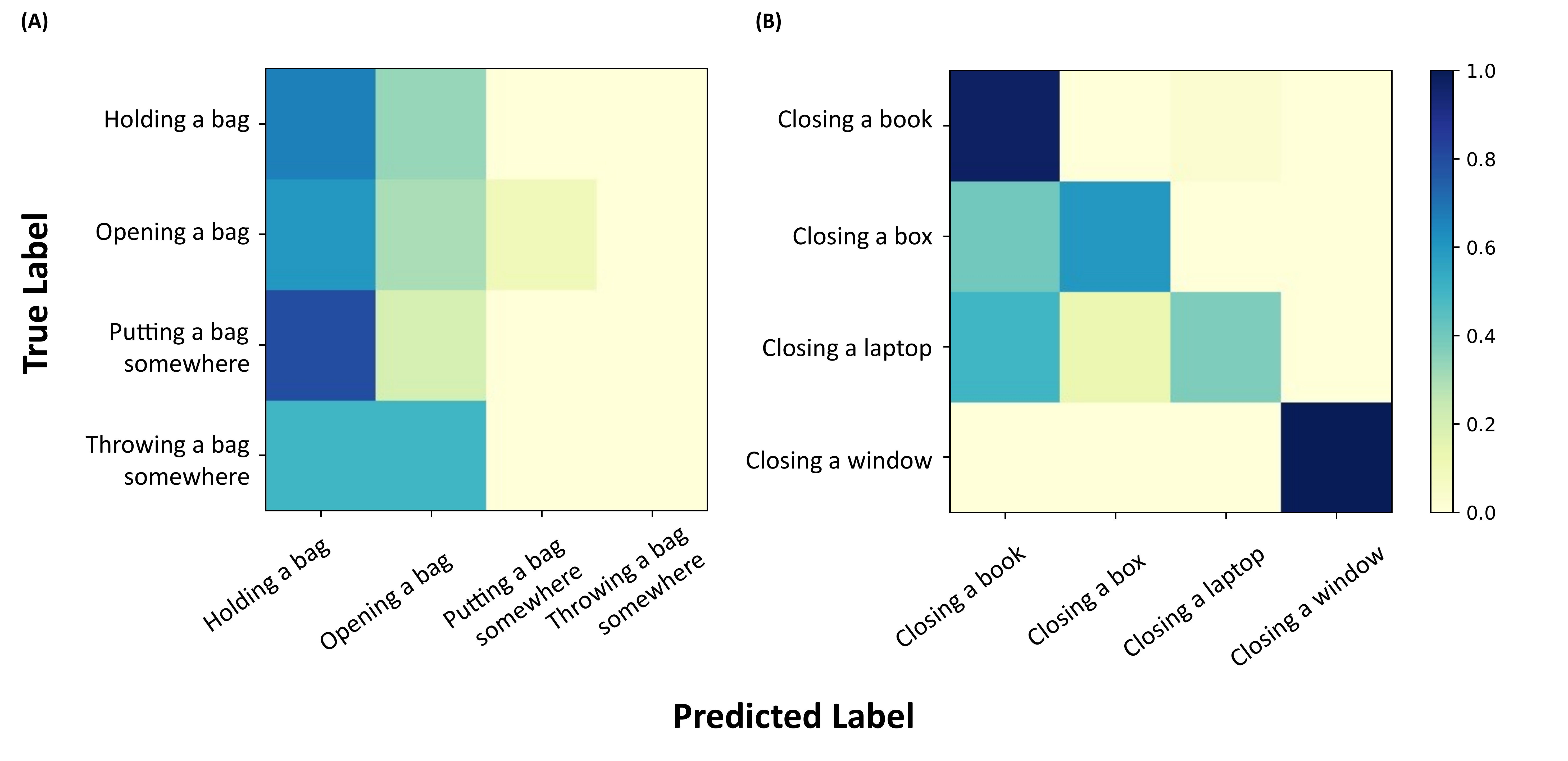}
\captionsetup{font=small} 
\caption{\textbf{Confusion matrix.} (A) Sub-matrix representing an object-cluster (e.g., 'bag'). (B) Sub-matrix representing a verb-cluster (e.g., 'closing'). For these examples, we binarized the predicted outputs with a threshold equal to 0.5 on the confidence value.}
\label{fig:confusion_matrix}
\end{figure}

The results reported in Table \ref{tab:metrics} clearly show that discriminating among verbs or objects poses differing levels of difficulty for the method. By comparing the mAP in the Object Seen (OS) scenarios, the method achieves higher performance when the verb is unknown (VU). Conversely, when the object is unknown (OU columns), the knowledge of the verb does not significantly enhance performance (VS-OU vs. VU-OU). Moreover, comparing scenarios where one of the elements is unknown (VU-OS vs. VS-OU), we observe a significantly higher mAP when object is seen, despite the comparable number of classes in the two scenarios (8 vs. 6).  However, in cases where the action does not involve an object (VU-NO), our algorithm demonstrates good generalization, outperforming the other conditions, especially the case VU-OS. This analysis may suggest that our method exhibits strong generalization to new verbs but in scenarios that require compositionality (i.e., discriminating between actions performed on different objects), spatial generalization dominates temporal generalization. It implies that objects play a crucial role in discrimination, as discriminating between actions like ``closing a book'' and ``closing a laptop'' is easier than distinguishing between ``opening a bag'' and ``holding a bag''. This hypothesis can be further supported by the results shown in Fig~\ref{fig:confusion_matrix}. We generated the confusion matrix for the unseen classes and conducted an analysis, grouping them based on objects or verbs. It is noteworthy that the method struggles in accurately identifying the verbs associated with the object (as shown in panel (A)), while it effectively exploits the object to distinguish classes with equal temporal dimensions.

\begin{table}[htbp]
    \centering
    \begin{tabular}{cccc}
        \toprule
        & OS & OU & NO \\
        \midrule
        VS & 16.41 (59) & 9.46 (6) & - \\
        VU & 25.21 (8) & 7.07 (1) & 27.47 (5)\\
        \bottomrule
    \end{tabular}   
    \captionsetup{font=small} 
    \caption{\textbf{mAP for different sub-sets of the testing set.} O-object, V- verb, S- seen, U- unseen, N- not present. The parenthesis contains the respective number of classes.}
    \label{tab:metrics}
\end{table}

\subsection{Spatial vs Temporal generalization}
Based on the findings discussed in the previous section, we wanted to explore a different training approach to assess whether the newly acquired prompt exhibits enhanced or distinct generalization characteristics in spatial versus temporal contexts. Table \ref{tab:map_diff_splits} shows interesting results. It clearly demonstrates that (1) the training protocol can impact the precision of the method, and (2) it reaffirms the advantages of exposing the model to more objects (object-augmentation) compared to more verbs (verb-augmentation). However, it remains unclear whether the focus is primarily on objects or verbs. As our intention is to deploy our module in a robotic application, achieving generalization across different objects is imperative.

\begin{table}[H]
    \centering
    \caption{mAP in zero-shot (\textit{ZSL}) and generalize zero-shot (\textit{GZSL}) settings for Verb vs Object splits.}
    \label{tab:example}
    \begin{tabularx}{0.4\textwidth}{XXXX} 
        \toprule
        \multicolumn{2}{c}{Verb Split} & \multicolumn{2}{c}{Object Split} \\
        \cmidrule(lr){1-2} \cmidrule(lr){3-4}
        \textit{ZSL} & \textit{GZSL}  & \textit{ZSL} & \textit{GZSL} \\
        \midrule
         17.06 & 21.62 & 6.84 & 15.90 \\
        \bottomrule
    \end{tabularx}
    \label{tab:map_diff_splits}
\end{table}

\section{CONCLUSIONS}
In this study, we introduced a novel unified framework, Dual-VCLIP, designed for zero-shot multi-label action recognition. Leveraging the robust vision-language pretraining from a large-scale dataset, our method addresses multi-label recognition tasks through the incorporation of a lightweight learnable overhead, that is a pair of positive and negative prompts along with the target class name as linguistic input. In addition, to better aggregate temporal visual frame features for each class, we reformulate the original visual attention in the pretraining model as a class-specific frame feature aggregation. Our promising results on the Charades dataset and the comparison with current SOTA methods and baselines demonstrate the effectiveness of our proposed approach. With our analysis on class compositionality, we showed that the learning protocols can play a significant role in the categorization process, especially when the method needs to generalize both in temporal and spatial dimensions, as needed for action recognition. This can suggest the importance of adopting conditional learning protocols, where the prompt is fine-tuned after learning verbs or objects. 
This aspect holds significant implications for zero-shot/few-shot scenarios in Human-Robot cooperative environments, where robots need to learn new tasks quickly and dynamically. Finding efficient zero-shot/few-shot learning protocols becomes crucial when limited data available for training the robots, when they may encounter new objects, actions, or scenarios that were not part of their initial training data and to use resources in more efficient way. Instead of collecting and labeling large amounts of data for every new task, robots can learn from a few examples or even from textual descriptions, reducing the time and effort required for training. This study also highlights the significance of introducing diagnostic tools and more comprehensive validation mechanisms when assessing these methodologies on intricate datasets such as Charades.

Finally, we acknowledge that this study opens up a number of questions that need further investigation. In the future, we plan to explore advancements to enhance the capabilities of our model. In particular, upcoming work will involve running experiments on multiple splits, validating over additional similar composed datasets, and exploring improvements through conditional learning, by integrating evidence debiasing constraints into the optimization objective to reduce static bias in video representations.

\addtolength{\textheight}{-9cm}   




\section*{ACKNOWLEDGMENT}
The authors would like to thank Francesco Brand for his contributions to the discussions on early design of the architecture. This work was supported by the Fit for Medical Robotics (Fit4MedRob) project - PNRR MUR Cod. PNC0000007 - CUP: B53C22006960001.

\printbibliography

\end{document}